%% file: main.tex
\documentclass[10pt,twocolumn,letterpaper]{article}

\usepackage{cvpr}
\usepackage{times}
\usepackage{epsfig}
\usepackage{graphicx}
\usepackage{amsmath}
\usepackage{amssymb}

\usepackage{helvet}  
\usepackage{courier}  
\usepackage{url}  

\usepackage{hyperref}
\usepackage{color}
\usepackage{booktabs}
\usepackage{multirow}
\usepackage{xspace}

\cvprfinalcopy 

\def\cvprPaperID{****} 

\begin{document}

\title{Efficient Fusion of Sparse and Complementary Convolutions}


\author{Chun-Fu (Richard) Chen, Quanfu Fan,  Marco Pistoia\\
IBM T.J. Watson Research Center, Yorktown Heights, NY 10598, USA \\
\tt\small{\{chenrich, qfan, pistoia\}@us.ibm.com}
\and
Gwo Giun (Chris) Lee \\
National Cheng Kung University, Tainan 70101, Taiwan \\
\tt\small{clee@mail.ncku.edu.tw}
}

\maketitle

\def\nonscConv{\textit{non\_scConv}\xspace}
\def\nonscFusion{\textit{non\_scFusion}\xspace}
\def\scConv{\textit{scConv}\xspace}
\def\scFusion{\textit{scFusion}\xspace}
\def\scFusionSingle{\textit{scFusion}$_{add}$\xspace}
\def\SC{\textit{SC}\xspace}
\def\etal{\emph{et al.}}
\definecolor{QF}{rgb}{0.7, 0.588, 0.178}

\maketitle
\begin{abstract}
We propose a new method to create compact convolutional neural networks (CNNs) by exploiting sparse convolutions. Different from previous works that learn sparsity in models, we directly employ hand-crafted kernels with regular sparse patterns, which result in the computational gain in practice without sophisticated and dedicated software or hardware. The core of our approach is an efficient network module that linearly combines sparse kernels to yield feature representations as strong as those from regular kernels. We integrate this module into various network architectures and demonstrate its effectiveness on three vision tasks, object classification, localization and detection.
For object classification and localization, our approach achieves comparable or better performance than several baselines and related works while providing lower computational costs with fewer parameters (on average, a $2-4\times$ reduction of convolutional parameters and computation). For object detection, our approach leads to a VGG-16-based Faster RCNN detector that is 12.4$\times$ smaller and about 3$\times$ faster than the baseline.


\end{abstract}

\input{tex/introduction}
\input{tex/relatedworks}
\input{tex/proposed}

\input{tex/experiments}
\input{tex/conclusion}

{\small
\bibliographystyle{ieee}
\bibliography{reference}
}

\clearpage
\newpage
\appendix
\input{tex/supplementary}
\end{document}

%% file: tex/introduction.tex
\section{Introduction}
\label{sec:intro}

Object recognition has made significant progress with convolutional neural networks (CNNs)~\cite{Krizhevsky_NIPS_2012,Simonyan_ICLR_2015,Huang_2017_CVPR}. To achieve competitive performance, CNNs need to increase their sizes substantially with deep architectures. 
Unfortunately, the increased complexity of these models 
makes it impractical to deploy them to real-time visual analysis or resource-constrained applications.

Recently there have been significant research efforts in making CNN models smaller and faster.
One research line aims to develop extremely compact network architectures 
for mobile devices with limited computational power, e.g., SqueezeNet~\cite{Iandola_ICLR_2017}, MobileNet~\cite{MobileNetV2} and ShuffleNet~\cite{ShuffleNetV2}.
Another direction focuses on model compression to reduce parameter redundancy in pre-trained models.
The techniques explored in the literature include low-rank approximation~\cite{Fan_BMVC_2017,Zhang_CVPR_2016}, kernel sparsification~\cite{Liu_CVPR_2015,Han_ICLR_2016} and network thinning~\cite{Liu_ICCV_2017,NISP}, just to name a few.
These approaches demonstrate ability to reduce model size and computational time greatly without compromising much on the classification accuracy. However, the sparsities learned in these models are often irregular, leaving the computational gain in practice either too small or highly dependent on dedicated software or hardware handlings~\cite{Liu_CVPR_2015,Han_ISCA_2016}.


\begin{table}[t!]
\centering
\caption{Comparison with related works.}
\label{table:analytic_comparison}
\resizebox{\linewidth}{!}{
\begin{tabular}{ccc}
\toprule
\multirow{2}{*}{Related works}  & \multirow{2}{*}{Key features} & Orthogonality\\
                                &                               & to \scFusion \\
\midrule
\cite{Jaderberg_BMVC_2014,Zhang_CVPR_2016,Liu_CVPR_2015,Yu_CVPR_2017}  & Low-rank approximation  & Y \\
\cite{Szegedy_CVPR_2016,Sun_ECCV_2016} & Low-rank kernels  & N \\
\midrule
\cite{Han_ICLR_2016,Liu_CVPR_2015} & Random sparsification  & Y \\
\cite{Wen_NIPS_2016,Yu_2017_CVPR,Fan_BMVC_2017,Peng_ECCV_2018} & Structural sparsification & Y \\
\midrule
\multirow{2}{*}{\cite{Ioannou_ICLR_2016,Iandola_ICLR_2017}} & Mixed-rank kernels with  & \multirow{2}{*}{Y} \\
     & channel-wise kernel fusion  & \\
\midrule
\cite{Molchanov_ICLR_2017,Liu_ICCV_2017,Luo_ICCV_2017,Hao_ICLR_2017,NISP}  & Network thinning  &   Y   \\
\midrule
Proposed & \scFusion  & $-$ \\
\bottomrule
\end{tabular}
} %
\end{table}

In this work, we propose an efficient approach based on the sparse kernel to reduce parameter redundancy in CNNs.
Unlike other model compression techniques that learn sparse representations, our approach exploits the hand-crafted spatial complementary (\SC) kernels. The \SC kernels, recently proposed by Fan~\emph{et al.}~\cite{Fan_BMVC_2017}, are paired sparse kernels that spatially complement each other by having zero weights either at the even or odd indices of a full kernel. 
A nice property of the \SC kernels is that, while each of them only sees part of the receptive field, a spatial join of them covers the entire field.
This allows for an effective combination of them to enhance feature representation for more complex kernels. 
The approach in~\cite{Fan_BMVC_2017} combines the \SC kernels by \textit{sequential filtering}, which alternatively stacks them at different layers within a model, in a similar spirit to separate filtering~\cite{Jaderberg_BMVC_2014}.
Nonetheless, the gain from such a combination is only about 1.5$\times$ reduction of convolutional parameters and computations.

To address such a limitation, we explore two ways in this paper to fuse the sparse kernels.
The first one, namely \textit{pairwise kernel fusion}, approximates a full convolution by fusing the paired \SC features in parallel.
As illustrated in Fig.~\ref{fig:scFusion}, such a method has the advantage over sequential filtering of covering all the pixels in the receptive field, potentially reserving better local image details in the feature representation, as validated in the object classification, localization and detection tasks in the experiments. Note that the computational overhead of the proposed fusion is negligible compared to the cost of convolution. The composite features, along with the \SC features themselves, are further joined together by a $1\times1$ convolution in the channel dimension, which is referred to as \textit{channel-wise kernel fusion}.
Based on these two types of kernel fusions, We design and implement a module that can be used to instantiate any popular CNN architecture. 
As shown later, the pairwise fusion can greatly strengthen the following channel-wise kernel fusion, thus resulting in powerful sparse feature representations with fewer base kernels and more compression.

Another motivation behind our design is that the deterministic sparse patterns in the \SC kernels make it easy to map them to various computing platforms without much effort, thus can result in high throughput in practice.
On the other hand, previous approaches based on sparsity learning such as~\cite{Liu_CVPR_2015,Han_ICLR_2016,Wen_NIPS_2016} could lead to performance degradation as more overheads are required to solve the irregularity in the learned sparse representation.

The contributions of our work are three-fold. 
First, we propose an efficient and powerful feature representation of CNNs and demonstrate its applicability on various CNN architectures.  
Second, we validate the superior adaptability our approach to several vision tasks including object classification, localization and detection through extensive experiments.
Finally, we show that our proposed approach achieves a compact and efficient Faster-RCNN detector, which is up to 3$\times$ faster and 12.4$\times$ smaller than the baseline while staying competitive in performance.

%% file: tex/relatedworks.tex
\section{Related Works}
\label{sec:relatedworks}

A great number of approaches have been proposed to reduce redundant parameters and computational complexity in CNN models. We highlight some works most relevant to ours in Table~\ref{table:analytic_comparison}. 
Below we provide a detailed review of the related works.

\textbf{Low-rank Approximation} factorizes learned kernels into low-rank kernels or directly deploys hand-crafted low-rank kernels.
Factorization-based approaches decompose high-rank kernels in a pre-trained model into multiple low-rank kernels~\cite{Zhang_CVPR_2016,Jaderberg_BMVC_2014,Liu_CVPR_2015,Yu_CVPR_2017}.
Decomposition techniques include principal component analysis, singular value decomposition, data-driven low-rank approximation or non-linear approximation.
On the other hand, hand-crafted low-rank kernels allow for training a network from scratch~\cite{Sun_ECCV_2016,Szegedy_CVPR_2016}. Approaches in this category sequentially cascade low-rank kernels to approximate a full kernel, for instance, stacking two 1-D kernels to approximate one 2-D kernel. However, the information lost in the low-rank kernels may be challenging for these approaches to recover. Our approach overcomes this issue by exploiting sparse and complementary kernels for feature representation.
Furthermore, most of these approaches are orthogonal to our work, so potentially we can apply them to further compress our models.

\textbf{Kernel Sparsification} learns sparsity by group regularization~\cite{Liu_CVPR_2015,Han_ICLR_2016,Wen_NIPS_2016,Peng_ECCV_2018}. Those approaches usually require a pre-trained model and then use a regularizer to obtain random or structural sparsity in the trained kernels. Re-training is needed for these approaches to recover performance. 
The sparsity is learned based on a random or structural weight regularizer, which forces a group of weights to be close to zeros, and then pruning those group of weights.
These approaches achieve high model compression rates on fully-connected layers which contain the most parameters of a model.
One big limitation of these approaches is that they usually require a dedicated computing platform due to random sparsity. This could lead to insignificant speedups. Furthermore, an overhead with random sparsity is that an additional table or a compressed representation is needed to record the non-zero positions. Our approach does not have such an issue as the sparse kernels used in our approach are highly structured, and implementation of such kernels is trivial.
Moreover, those approaches can be deployed on top of our approach to have an even more compact model.

\textbf{Mixed-rank Kernels with Channel-wise Kernel Fusion} groups kernels of different ranks in the channel dimension. 
Ioannou~\etal~\cite{Ioannou_ICLR_2016} propose to mix two 1-D kernels ($-$ and $\vert$ shape) and one optional 2-D kernel for feature representation. Nonetheless, they use non-complementary kernels which are too sparse to extend to large kernel size. In ~\cite{Iandola_ICLR_2017}, Iandola~\etal~further propose to mix $1\times1$ and $3\times3$ kernels together and combine them by a $1\times1$ filter. Our approach enhances their work by fusing sparse and complementary kernels to enrich feature representations. 

\textbf{Network Thinning} is an approach to remove multiple kernels of a convolutional layer to thin a CNN model. This approach uses a pre-trained model with a large amount of data to measure the importance of each kernel and discarding unimportant kernels; then, a re-training is required to recover its performance~\cite{Molchanov_ICLR_2017,Liu_ICCV_2017,Luo_ICCV_2017,Hao_ICLR_2017,NISP}. Our approach can be beneficial with this approach to further compress our model since the slimming approach is applicable to any trained models.


More recently there are some attempts to explore model compression through knowledge distillation~\cite{Hinton_arXiv_2015}. While this is a promising direction, most of the works have not been tested on large datasets such ImageNet yet.

%% file: tex/proposed.tex
\section{Our Approach}
\label{sec:ourapproach}

In this section, we develop a compact module based on the \SC kernels
proposed in~\cite{Fan_BMVC_2017}. Different from Fan's work that stacks the sparse kernels sequentially, our approach fuses them pairwisely to form a more efficient and stronger feature representation. 
We first introduce the \SC kernels briefly and then detail our approach below.

\subsection{Base Sparse Kernels}
\label{subsec:ourapproach:base}
\begin{figure}[!b]
    \centering
    \includegraphics[width=0.65\linewidth]{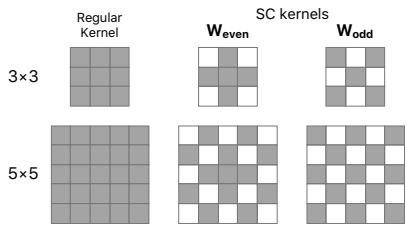}
    \caption{3$\times$3 and 5$\times$5 regular and \SC kernels. White color indicates zero weights here.}
    \label{fig:sckernel}
\end{figure}

\SC kernels are defined as a pair of sparse kernels that spatially complement each other. The base kernels used in~\cite{Fan_BMVC_2017} are the simplest form with regular sparse patterns. One of the kernels, these two kernels, denoted by denoted by $\mathbf{W_{even}}$ and $\mathbf{W_{odd}}$ respectively, have zero weights either
at the even or odd indices of a full kernel. Apparently, \SC kernels can be extended to any kernel size larger than 1. Fig.~\ref{fig:sckernel} illustrates an example of $3 \times 3$ and $5\times5$ kernels. Note that the center point of $\mathbf{W_{even}}$ is not zeroed out as this location often carries a large weight, which is better not overlooked.


\SC kernels are deterministic, meaning that there is no need to use an index table to store their sparse patterns in memory. This is a big advantage over many other sparse representation such as~\cite{Liu_CVPR_2015,Wen_NIPS_2016,Han_ICLR_2016}. Such a simple form also provides the regularity in computation, enabling most of computing platforms to increase the throughput without much effort, as demonstrated in the experiment section.


The $1\times3$ and $3\times1$ kernels used in~\cite{Ioannou_ICLR_2016} can be considered as a special type of deterministic sparse kernels, but they are not spatially complementary. Dilated convolution~\cite{Yu_CVPR_2017} is another type of sparse representation, which expands the kernel size by filling the empty positions with zeros. Hence it is structurally different from \SC kernels. Dilated convolution is usually used to approximate large-size kernels, and mostly applied to semantics segmentation~\cite{Yu_2017_CVPR}.

\subsection{Fusion of \SC Kernels}
\label{subsec:ourapproach:fusion}

\begin{figure}[!b]
    \centering
    \includegraphics[width=0.8\linewidth]{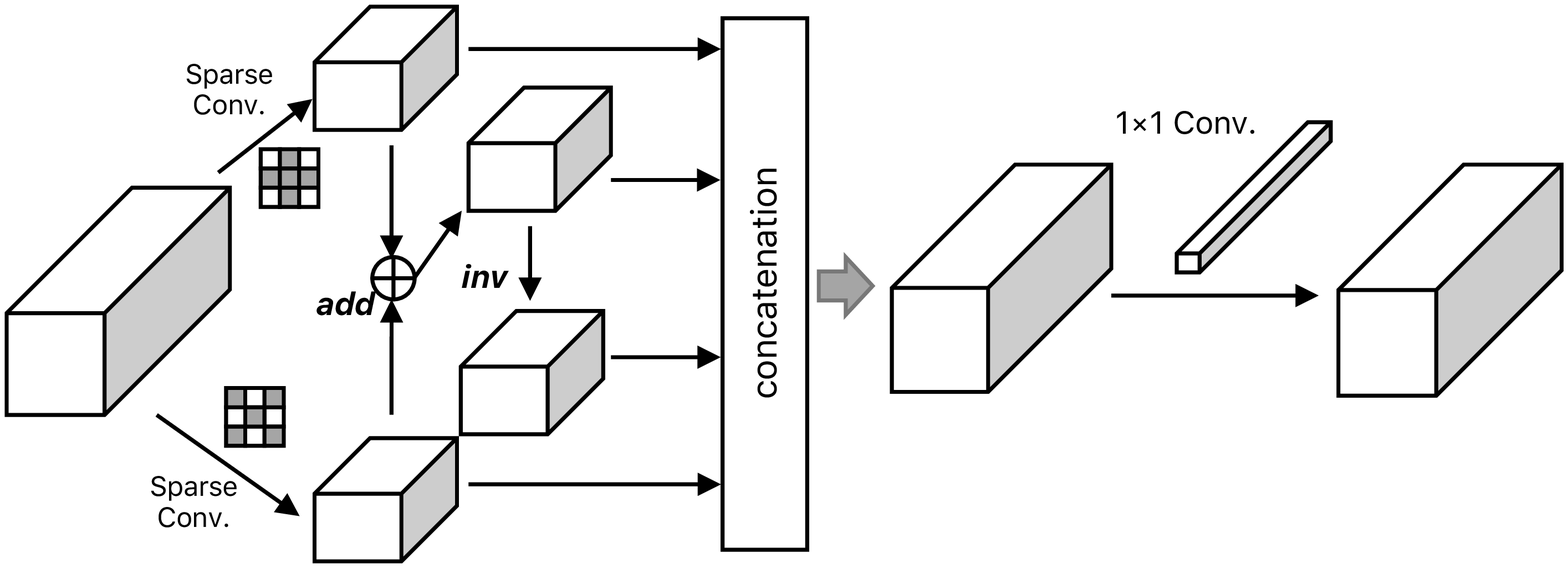}
    \caption{Our proposed module (\scFusion). Two sparse base kernels with complementary patterns are applied in parallel to extract features, which are joined together with two linear operations, \textit{addition} and \textit{inverse}. A subsequent $1\times1$ filter further linearly combines the sparse features as well as the joint features. The white color in the sparse kernels denotes the zero weights.}
    \label{fig:scFusion}
\end{figure}

Sequential filtering stacks two or more small kernels in sequence to represent a larger one.
This idea has proven effective for decreasing the number of parameters and computations in CNN models such as VGG~\cite{Simonyan_ICLR_2015}. 
Recently, channel-wise kernel fusion has been used in approaches like~\cite{Iandola_ICLR_2017,Ioannou_ICLR_2016} to group kernels of different shapes across channels. 
To take advantage of the complement property of our base kernel, we propose to join them spatially at the same convolutional layer to approximate the response of a full kernel.

First, we apply the \SC kernels with the input in parallel to extract the features in different phases.
We then combine the \SC kernels pairwisely by a linear \textit{addition} 
 (\textit{pairwise kernel fusion}) to cover the full receptive field.
The responses of the \SC kernels, together with the fused responses, are subsequently concatenated and further combined by an $1\times1$ filter to learn a more discriminative feature representation (\textit{channel-wise kernel fusion}). Note that we introduce non-linearity before the channel-wise kernel fusion by ReLU. 
Our idea is illustrated in Fig.~\ref{fig:scFusion}. Pairwise kernel fusion itself does not bring any savings of FLOPs or model parameters. Nonetheless, as shown later in the experiment section, the two sparse kernels, when combined together, enable our approach to learn powerful sparse models with fewer base kernels, leading to less computation as well as fewer parameters in the models. 



To further enrich the feature representation in our approach, we apply the \textit{inverse} operation proposed in~\cite{Shang_ICML_2016} to generate symmetric feature maps from the fused responses of the \SC kernels.
The \textit{inverse} operation is a unary operation, which negates the responses of a kernel before a ReLU layer. It is mathematically equivalent to adding to the model a kernel $\mathbf{W}_i$ for each pair of \SC kernels where $\mathbf{W}_i = - (\mathbf{W_{even}}_i + \mathbf{W_{odd}}_i)$, and $i$ is the $i^{th}$ base kernel in a layer. 
The \textit{inverse} fusion incurs almost no computation compared to the cost of full convolutions while greatly strengthening the fused sparse feature representation. As shown later in the experiments, such enhancement is critical for our approach to achieve good accuracy in scenarios when fewer base kernels are desired for more computational savings.



\begin{figure}[!b]
    \centering
    \includegraphics[width=0.7\linewidth]{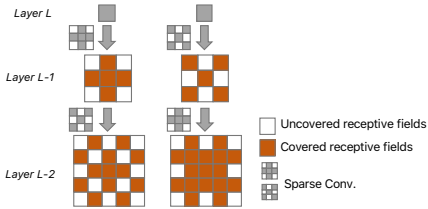}
    \caption{Receptive fields covered by the \SC kernels with sequential filtering. Not all pixels can be covered regardless of how the kernels are combined.}
    \label{fig:receptiveFields}
\end{figure}

There are a few advantages of our approach over the sequential filtering method adopted in~\cite{Fan_BMVC_2017}. Firstly, \scFusion is more flexible, and can be integrated into any popular network architectures to reduce computation and model parameters, as demonstrated in the experiment section. Secondly, \scFusion covers the entire receptive field without overlooking any pixel like a regular kernel does, thanks to the nice spatially complementary property of the \SC kernels. This potentially benefit vision tasks such object localization and detection, which requires local image details in the feature representation. On the other hand, alternatively stacking the \SC kernels always results in receptive fields
incompletely covered, as shown in Fig.~\ref{fig:receptiveFields}.



\subsection{Complexity Analysis}
\label{subsec:ourapproach:complexity}

\begin{table}[t]
\centering
\caption{Complexity reduction ratio $\rho$ of \scFusion.}
\label{table:complexity-reduction}
\resizebox{0.7\linewidth}{!}{
\begin{tabular}{c|ccc}
\toprule
  $k=3$  &   $\alpha=2$  &   $\alpha=4$  &   $\alpha=8$  \\
\midrule
$C_{out}/C_{in}= 1$ & 1.29$\times$&   2.57$\times$&   5.14$\times$  \\
$C_{out}/C_{in}=2$ &   1$\times$   &   2$\times$   &   4$\times$  \\
\bottomrule
\end{tabular}
} %
\end{table}

The algorithmic complexity of \scFusion is directly controlled by the number of base sparse kernels. Consider a convolutional layer with an input tensor of $C_{in}\times H\times W$. Here $C_{in}$ is the number of input kernels and $(H,W)$ is the dimension of a feature map. Then the total computation of the layer is $k^2C_{in} C_{out} H W$ where $C_{out}$ is the number of output kernels and $k$ is the kernel size. If the same dimension of input and output tensors are maintained while using $n$ sparse base kernel each in \scFusion, it will require $2\times \lceil k^2/2 \rceil C_{in} n H W + 4 n C_{out} H W$, the second term of which comes from the channel-wise fusion (The input channels of $1\times1$ convolution are four times as large as the number of kernels of one sparse convolution.). Hence, the reduction ratio $\rho$ of FLOPs can be expressed by
\begin{equation}
    \begin{split}
    \label{eqn:reductionratio}
        \rho = & \frac{k^2C_{in} C_{out} H W}{2\times \lceil k^2/2 \rceil C_{in} n H W + 4 n C_{out} H W} \\
               = & \frac{k^2 C_{out}}{2\times \lceil k^2/2 \rceil n + 4 n \frac{C_{out}}{C_{in}}} 
               = \frac{\alpha k^2}{ (2 \times \lceil k^2/2 \rceil + 4\frac{C_{out}}{C_{in}})}
    \end{split}
\end{equation}


We further define a complexity controlling parameter $\alpha = C_{out}/n$, which indicates the ratio of the number of kernels of a baseline model over the number of kernels for each base sparse kernel of \scFusion. Thus, $\rho$ can be represented as a function of $\alpha$. A larger $\alpha$ suggests a more compact and more efficient \scFusion model.
The reduction ratio of model parameters can be derived in the same way.

Table~\ref{table:complexity-reduction} shows the reduction ratios of FLOPs and parameters under different $\alpha$ and different ratios of output size over input size ($C_{out}/C_{in}$). A typical setting of a CNN model in our experiments is $C_{out}/C_{in}=1$ for the layers without spatial downsampling and $C_{out}/C_{in}=2$ for those with spatial downsampling. Under such a setting, $\alpha=4$ results in a  $\rho$ of $2.57\times$ and $\alpha=8$ leads to a larger $\rho$ of $5.14\times$.



%% file: tex/experiments.tex
\section{Experimental Results}
\label{sec:experiment}
We conducted extensive experiments below to validate the effectiveness of our proposed approach on object classification, and compared our models with other works of sparse modeling.
We also demonstrated the adaption capability of our approach on another two vision tasks: object localization and object detection.

\input{tex/exp-classification.tex}
\input{tex/exp-localization.tex}

\input{tex/exp-detection.tex}

%% file: tex/exp-classification.tex
\subsection{Object Classification}
\subsubsection{Experimental Setup}
We first evaluated our proposed approach on classification tasks using CIFAR-10~\cite{cifar10} and ImageNet~\cite{ILSVRC15}.
A \scFusion version of a reference CNN model is created by replacing all convolutional layers with the \scFusion modules described in the Our Approach section.
The number of input and output channels in a \scFusion module is set to the exactly the same as that of the reference model for fair comparison. 
We chose the Caffe deep learning framework for model training and test~\cite{Jia_ACMMM_2014}.

We trained all the models from scratch using the identical hyper-parameters of the baseline networks, and all training details can be found in the supplementary materials.

\begin{table}[bt]
\centering
\caption{Classification accuracy on CIFAR-10.}
\label{table:cifar10-classification}
\resizebox{\linewidth}{!}{
\begin{tabular}{l|ccccc}
\toprule
& \multirow{2}{*}{Model} & \multirow{2}{*}{Top-1 Acc.} & \multirow{2}{*}{FLOPs $\downarrow$} & Params $\downarrow$ \\
&                          &           &              &   (Total) \\

\midrule
\multirow{3}{*}{{ResNet-32}~\cite{He_CVPR_2016}} &baseline     & 93.98\% & $-$ & $-$\\
&\scFusion-2 & \textbf{94.37\%} & 1.26$\times$ &1.25$\times$  \\
&\scFusion-4 & 93.16\% & 2.45$\times$ & 2.39$\times$  \\
\midrule
\multirow{2}{*}{{ResNet-164}~\cite{He_2016_ECCV}} &baseline     & 94.58\% & $-$ & $-$\\
&\scFusion-2 & \textbf{95.21\%} & 1.28$\times$ & 1.27$\times$  \\
&\scFusion-4 & 94.89\% & 2.56$\times$ & 2.51$\times$  \\
\midrule
\multirow{2}{*}{{DenseNet-40}~\cite{Huang_2017_CVPR}}& baseline   & 93.89\% & $-$ & $-$\\
&\scFusion-2 & \textbf{95.01\%} & 1.53$\times$ & 1.60$\times$  \\
&\scFusion-4 & 94.60\% & 2.37$\times$ & 2.62$\times$  \\
\bottomrule
\end{tabular}
} 
\end{table}


\begin{figure}[bt]
    \centering
    \includegraphics[width=0.8\linewidth]{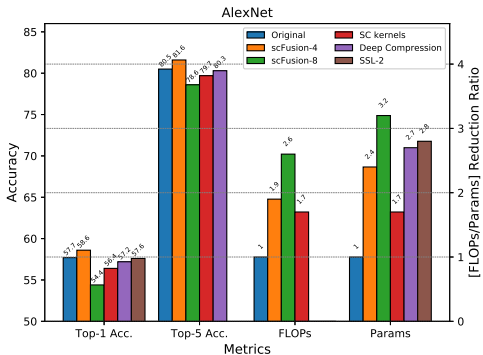}
    \caption{[AlexNet] Performance on ImageNet of \scFusion and other approaches.}
    \label{fig:comp-alexnet}
\end{figure}


For clarity, a model studied below is named as \scFusion-$\alpha$ where $\alpha \in \{2, 4, 8\}$ is the reciprocal of the complexity controlling parameter defined in the previous section.
We applied the same $\alpha$ for all the convolutional layers in our experiments. In what follows, the experimental results are reported using top-1 and top-5 accuracies as well as reductions in FLOPs and parameters.

\subsubsection{Results on CIFAR-10}
On CIFAR-10, we tested our approach with several CNNs such as ResNet~\cite{He_CVPR_2016,He_2016_ECCV} and DenseNet~\cite{Huang_2017_CVPR}

Table~\ref{table:cifar10-classification} lists the results of different CNNs. Our approach performs comparably against or better than the baseline models under different architectures while providing over $2\times$ savings of model parameters and FLOPs (\scFusion-4).
This clearly demonstrates that \scFusion is capable of providing not only efficient but also effective feature representations for object classification.

\begin{figure}[bt]
    \centering
    \includegraphics[width=0.8\linewidth]{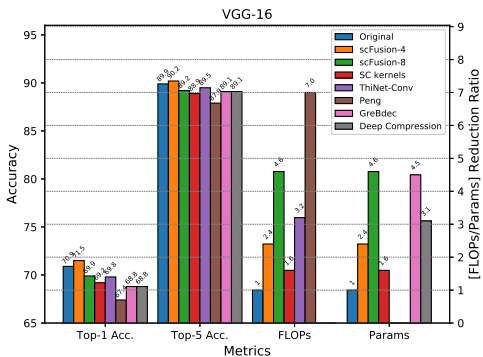}
    \caption{[VGG-16] Performance on ImageNet of \scFusion and other approaches.}
    \label{fig:comp-vgg16}
\end{figure}

\begin{figure}[bt]
    \centering
    \includegraphics[width=0.8\linewidth]{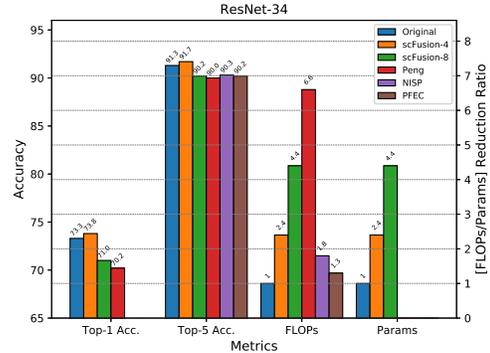}
    \caption{[ResNet-34] Performance on ImageNet of \scFusion and other approaches.}
    \label{fig:comp-resnet34}
\end{figure}

\subsubsection{Results on ImageNet}
\label{subsubsec:exp-imagenet}

On ImageNet, we evaluated our proposed approach with some popular CNNs including AlexNet~\cite{Krizhevsky_NIPS_2012}, VGG-16~\cite{Simonyan_ICLR_2015} and ResNet-34~\cite{He_CVPR_2016}.
In all these networks, we keep the first convolutional layer dense since sparsifying the first convolutional layer leads to a slight performance drop; furthermore, the first layer only accounts for a small portion of the parameters and computations, we keep it intact in our models. 
Both $3\times3$ and $5\times5$ kernels are sparsified in these models, suggesting that \scFusion is effective regardless of kernel size.

The classification results of the models are summarized in Fig.~\ref{fig:comp-alexnet},~\ref{fig:comp-vgg16} and~\ref{fig:comp-resnet34}.
Since \scFusion only compresses the convolutional layers, we only show the FLOP and parameter reduction at convolutional layers on the figures.
Overall our approach demonstrates competitive performance against the AlexNet, VGG-16 and ResNet-34 baselines.
\scFusion-4 yields better top-1 and top-5 accuracies than the baselines while giving reasonable savings on model parameters and FLOPs.
On the other hand, when fewer base kernels (e.g. $\alpha=8$) are used, \scFusion-8 gains a significant reduction of over 4$\times$ in FLOPs and over 4$\times$ in convolutional parameters, with only about $1\%$ drop on top-1 accuracy for VGG-16.


\subsubsection{Comparison With Other Approaches}
We compared our approach with some recently developed techniques of model compression based on sparse kernel representation, including \SC Kernel~\cite{Fan_BMVC_2017}, Deep Compression~\cite{Han_ICLR_2016}, SSL-2~\cite{Wen_NIPS_2016}, ThiNet-Conv~\cite{Luo_ICCV_2017}, GreBdec~\cite{Yu_CVPR_2017}, NISP~\cite{NISP}, PFEC~\cite{Hao_ICLR_2017}, and Peng's work~\cite{Peng_ECCV_2018}. Note that if the results were not provided with some approaches among the above, they are left blank in the figures.


Among them, the work of deep compression~\cite{Han_ICLR_2016} is based on sparsity pruning while other approaches all learn sparsity in kernels via group regularization~\cite{Wen_NIPS_2016,Peng_ECCV_2018}. 
We also include in our evaluation two methods based on network thinning, ThiNet~\cite{Luo_ICCV_2017}, NISP~\cite{NISP}, PFEC~\cite{Hao_ICLR_2017}. These approaches focus on filter-level pruning, which removes an entire filter if it's insignificant.

As indicated by Fig.~\ref{fig:comp-alexnet},~\ref{fig:comp-vgg16} and~\ref{fig:comp-resnet34}, our approach outperforms all others while delivering comparable or better savings in terms of convolutional parameters and computations across all AlexNet, VGG-16 and ResNet-34. 
First, the notable performance gap between \scFusion and \SC kernels~\cite{Fan_BMVC_2017} strongly suggests that our fusion methodology is superior to sequential filtering. Compared to other approaches, \scFusion-8 achieves a much better balance between FLOPs reduction and accuracy on VGG-16 and ResNet-34 (Fig.~\ref{fig:comp-vgg16} and Fig.~\ref{fig:comp-resnet34}). 
On the other hand, Peng's approach~\cite{Peng_ECCV_2018} demonstrates a much larger reduction of FLOPs than our approach, but it results in a significant drop on the top-1 accuracy (2.5\% point on VGG-16 and 0.8\% point on ResNet-34) as compared to \scFusion-8. Nevertheless, as shown in Table~\ref{table:speedup-gpu}, the high theoretic FLOPs reductions achieved by this approach can only be limitedly realized on a GPU in practice. 


Approaches like \cite{Han_ICLR_2016,Yu_CVPR_2017} perform sparsification on the entire network architecture, so they benefit greatly from pruning the FC layers, which leads to a much higher reduction rate with regards to the total model size. Since our approach is orthogonal to many other model compression techniques, we applied the deep compression~\cite{Han_ICLR_2016} to further compress the FC layers in our VGG-16-based \scFusion-8 model.
As shown in Table~\ref{table:imagenet-compression-comparison}, this gives rise to a total reduction of 13.45$\times$, with only a slight drop in accuracy. Thus, our \scFusion-8 achieved compression ratio to others with competitive performance and the deterministic sparse patterns assists in improving the speed in practice.


\begin{table}[!t]
\centering
\caption{Comparison with other approaches on ImageNet with VGG-16}
\label{table:imagenet-compression-comparison}
\resizebox{\linewidth}{!}{
\begin{tabular}{l|cc}
\toprule
Network & Top-1 Acc. & Params$\downarrow$ \\
\midrule
\scFusion-8 with Deep Compression (ours)$^\dagger$  &   68.83\% &   13.45$\times$ \\
Deep Compression~\cite{Han_ICLR_2016}    &   68.83\%  &    13$\times^*$  \\
GreBdec~\cite{Yu_CVPR_2017} &   68.75\%  &   14.22$\times$   \\
\bottomrule
\multicolumn{3}{l}{\footnotesize{$^*$: The number of parameters are pruned.}} \\
\multicolumn{3}{l}{\footnotesize{$^\dagger$: Apply deep compression on \textbf{only} FC layers.}} \\
\end{tabular}
} 
\end{table}

\begin{figure}[bt]
    \centering
    \includegraphics[width=0.65\linewidth]{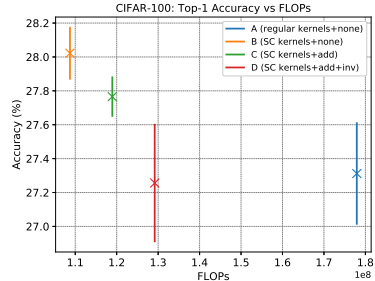}
    \caption{Ablation Study of \scFusion on CIFAR-100 with VGG-16. (Five runs of each configuration were conducted and the mean values and standard deviations were reported here.) A: \textit{regular} kernels without \textit{addition} and \textit{inverse};
    B: \SC kernels without \textit{addition} and \textit{inverse};
    C: \SC kernels with \textit{addition}; .
    D: \SC kernels with both \textit{addition} and \textit{inverse}.
    }
    \label{fig:scfusion-ablation}
\end{figure}


\subsubsection{GPU Speedup in Practice}

\begin{table}[bt]
\centering
\caption{Comparison of GPU speedup (image size: 224$\times$224).}
\label{table:speedup-gpu}
\resizebox{\linewidth}{!}{
\begin{tabular}{ll|ccc}
\toprule
\multicolumn{2}{l}{Network} & Top-1 Acc. & FLOPs$\downarrow$ & GPU speedup$\uparrow$ \\
\midrule
\multirow{2}{*}{VGG-16} & Peng~\cite{Peng_ECCV_2018} & 67.45\% & \textbf{7.04$\times$} & 1.34$\times$ \\
& \scFusion-8 (Ours)        & \textbf{69.90\%} & 4.64$\times$ & \textbf{2.02$\times$} \\
\midrule
\multirow{2}{*}{ResNet-34} & Peng~\cite{Peng_ECCV_2018} & 70.24\% & \textbf{6.60$\times$} & 1.21$\times$ \\
& \scFusion-8 (Ours) & \textbf{71.04\%} & 4.40$\times$ & \textbf{1.84$\times$} \\
\bottomrule
\end{tabular}
} 
\end{table}
One advantage of our approach is that the \SC kernels are deterministic sparse patterns. Nvidia's CuDNN accelerates convolutions by the Winograd algorithm~\cite{Lavin_CVPR_2016}. Modifying such an algorithm in \scFusion is trivial, only requiring us to skip the zero weights in the \SC kernels rather than coming up with a sophisticated and dedicated design for the irregular sparse patterns of a kernel from the approaches such as~\cite{Liu_CVPR_2015,Han_ISCA_2016}. We implemented this for \scFusion and conducted some preliminary benchmarking on a K80 GPU under the Caffe framework.  

As shown in Table~\ref{table:speedup-gpu}, our approach achieves a real speedup by nearly 2 times on both VGG-16 and ResNet-34, which is quite significant. As a comparison, we ran performance analysis using the models from Peng's work~\cite{Peng_ECCV_2018} under the same Caffe framework. As can be seen from Table~\ref{table:speedup-gpu}, these models only result in moderate speedups of 1.2-1.3, though they indicate significantly higher theoretic FLOPs reductions.
This is largely because Peng's approach intensively uses group convolutions to reduce FLOPs, which greatly limits the GPU's capacity in parallel computing and increases the memory access costs. 
Such a limitation was also observed in~\cite{ShuffleNetV2}, and discussed in detail in their paper (Guideline 2).

\subsubsection{Ablation Study of \scFusion}
To understand the contribution of each component in \scFusion, we ran a control experiment on CIFAR-100 over the VGG-16 network. Using the same module in Fig~\ref{fig:scFusion}, we started with the \SC kernels only, but without \textit{addition} and \textit{inverse}, indicated by Conf. B in Fig.~\ref{fig:scfusion-ablation}. We then added fusion (Conf. C) and \textit{inverse} (Conf. D), respectively. As a baseline, we also compared the regular kernels with the \SC kernels (Conf. B). We trained 5 models for each configuration and averaged the results. The contribution of each component of \scFusion is clearly seen in Fig.~\ref{fig:scfusion-ablation}. The regular kernels yield comparable results with \scFusion, but are significantly heavier in parameters.

%% file: tex/exp-localization.tex
\subsection{Object Localization}
\label{subsec:localization}

To demonstrate the adaption capability of our approach to other vision tasks, we evaluated our approach on the object localization task in this section.
We follow the methodology used in~\cite{Zhou_CVPR_2016} to extend the VGG-16-based \scFusion models (\scFusion-$\alpha$-GAP models) and use class attention map (CAM) for object localization (See the supplementary materials for details.). Shown in Table~\ref{table:localization-error} are the top-5 localization accuracy on the test set of ILSVRC-2014.
Note that the localization models achieve similar FLOPs reduction ratio to baselines to the classification models.


\begin{table}[t]
\centering
\caption{Localization accuracy on the ILSVRC-2014 test set.}
\label{table:localization-error}
\resizebox{0.8\linewidth}{!}{
\begin{tabular}{l|cc}
\toprule
Network                &  Top-5 Acc. \\
\midrule
VGG-16-\scFusion-4-GAP &\textbf{62.7\%} \\
VGG-16-\scFusion-8-GAP &  58.2\%  \\
GoogLeNet~\cite{Zhou_CVPR_2016} &   57.1\% \\
Backprop~\cite{Simonyan_ICLRWS_2014} &  53.6\% \\
\bottomrule
\end{tabular}
} 
\vspace{-5pt}
\end{table}

As indicated by Table~\ref{table:localization-error}, both \scFusion-4-GAP and \scFusion-8-GAP outperform the baseline by a large margin.
Our results are also better than GoogLeNet, the best results reported in~\cite{Zhou_CVPR_2016}. The localization performance of our approach seems to suggest that \scFusion can preserve local image details well, which is confirmed in the object detection task later.
On the other hand, our approach does object localization in a weakly supervised way, but still outperforms Backprop~\cite{Simonyan_ICLRWS_2014}, a fully supervised approach for the localization task.

%% file: tex/exp-detection.tex
\subsection{Object Detection}
\label{subsec:deetection}
\begin{table}[tb]
\centering
\caption{Object detection results on the VOC2007 test set.}
\label{table:detection-accuracy}
\resizebox{\linewidth}{!}{
\begin{tabular}{cl|c}
\toprule
&Detector          &  Accuracy (mAP) \\
\midrule
\multirow{5}{*}{\rotatebox{90}{AlexNet}} & baseline         & 62.2 \\
&\scFusion-4                                                & \textbf{62.3}  \\
&\SC kernels~\cite{Fan_BMVC_2017}                                & 61.1\\
&Deep Compression~\cite{Han_ICLR_2016}                                  & 54.3\\
&SSL-2~\cite{Wen_NIPS_2016}                                 & 58.5\\
\midrule
\multirow{5}{*}{\rotatebox{90}{VGG-16}} & baseline~\cite{Ren_TPAMI_2017}         & 73.2 \\
&\scFusion-8 (ours)                                                      & \textbf{73.9}  \\
&\scFusion-8-GAP (ours)                                                  & 40.9  \\
&\SC kernels~\cite{Fan_BMVC_2017}                                            &73.1\\
&ThiNet-Conv-GAP~\cite{Luo_ICCV_2017}                                           &34.2\\
\bottomrule
\end{tabular}
} 
\vspace{-8pt}
\end{table}

\begin{table}[tb]
\centering
\caption{Object detection results on COCO test2015-dev.}
\label{table:detection-coco}
\resizebox{0.7\linewidth}{!}{
\begin{tabular}{l|cc}
\toprule
Detector & \multicolumn{2}{c}{Ave. Precision IoU} \\
& 0.5:0.95 & 0.5 \\
\midrule
VGG-16~\cite{Ren_TPAMI_2017} & 21.9 & 42.7  \\ 
\scFusion-8 (ours)  & \textbf{22.8} & \textbf{43.3}  \\ 
\bottomrule
\end{tabular}
} %
\vspace{-8pt}
\end{table}


In this section, we validated our approach on the object detection task.
We used our \scFusion models as feature extractors and trained Faster-RCNN~\cite{Ren_TPAMI_2017} detectors on the PASCAL VOC~\cite{Everingham2010} and MS COCO dataset~\cite{Lin_2014_ECCV}. For PASCAL VOC, we combined the training and validation sets of VOC2007 and VOC2012 as training data, and then evaluated the detector on the VOC2007 test set. For MS COCO, we used the 2014 train + val35k as suggested in~\cite{Ren_TPAMI_2017} for training and test2015-dev for testing. 

\subsubsection{Detection Results on VOC2007 and MS COCO} We include 4 other model compression approaches in our evaluation, including Deep Compression~\cite{Han_ICLR_2016}, SSL-2~\cite{Wen_NIPS_2016}, ThiNet~\cite{Luo_ICCV_2017}, and \SC kernels~\cite{Fan_BMVC_2017}. They were also used in the evaluation of the object classification task.
When training Faster-RCNN with these models, we disabled updating of any zero weight from them except ThiNet~\cite{Luo_ICCV_2017} to maintain their original sparsity. With ThiNet, we append the classification and regression layers in Faster-RCNN directly to the global average pooling (GAP) layer of the classification model. For fair comparison, all detectors were trained and tested at the default settings given by~\cite{py-faster-rcnn}. 

Table~\ref{table:detection-accuracy} lists the results of object detection from all the approaches on the VOC2007 test set. Our approach is slightly better than the baseline while outperforms all the others in comparison.
Note that ThiNet~\cite{Luo_ICCV_2017} yields an extremely low accuracy on detection. We speculate that this has to do with the change in this model which substitutes the FC layers by a GAP layer. To confirm this, we trained a Faster-RCNN detector with our \scFusion-8-GAP model, which undergoes a similar structural modification. While doing better, this detector only leads to an accuracy of 40.9, which is still far beyond satisfactory. 

Table~\ref{table:detection-coco} shows the detection results on the more challenging MS COCO dataset. Again, our approach achieves better performance than the VGG-16 baseline.

\subsubsection{Efficient and Compact Detectors} We benchmarked our \scFusion-based detector on a machine with a Nvidia K80 GPU. As shown in Table~\ref{table:detection-speed}, our approach yields a speedup of $\sim$1.4$\times$ in computation. To further improve its efficiency, we trained a detector with multiple scales with the same image height and width [256$\times$256, 512$\times$512, 768$\times$768] and tested the detector on a scale of 512$\times$512 rather than the default 600$\times$1000 used in single scale training in Faster-RCNN. With this, the detector sees a significant speedup of more than 2$\times$ while achieving a better accuracy than the model trained on a single scale and larger resolution.

Since \scFusion is only applied to convolution layers, a remaining issue for a Faster-RCNN detector based on our approach is the size of the model, which is dominated by the parameters from the FC layers. To build a more compact detector, we applied sparse ROI pooling and reduced the number of nodes on the FC layers from 4096 to 512, similar to what's done in~\cite{Fan_BMVC_2017}. As can be seen from Table~\ref{table:detection-speed}, doing so leads to a very compact model as small as 42M, which sees another 1$\times$ speedup over the \scFusion-8 model trained with multiple scales. Overall, this detector is 12.4$\times$ smaller and 3$\times$ faster in inference than the VGG-16-based Faster-RCNN while staying competitive in performance. We expect that such well-performed but tiny detectors have important applications to resource-constrained devices such as mobile phones and FPGAs.

\begin{table}[tb]
\centering
\caption{Object detection results on VOC2007 test set with VGG-16-based Faster-RCNN.}
\label{table:detection-speed}
\resizebox{\linewidth}{!}{
\begin{tabular}{l|ccc}
\toprule
Detector     & Accuracy & Model & Speed \\
     & (mAP) & (MB) & (ms) \\
\midrule
baseline~\cite{Ren_TPAMI_2017}, single scale  & 73.2 & 522 & 275\\
\scFusion-8, single scale & 73.9 & 475 & 202 \\
\scFusion-8, multi-scale        & \textbf{74.4} & 475 & 126\\
\scFusion-8 (sp2-fc512), multi-scale    & 73.4 & \textbf{42} & \textbf{93}\\
\bottomrule
\multicolumn{4}{l}{\footnotesize{sp2: sparse ROI pooling with even-indexed features. fc512: the size of FC layer is reduced to 512.}} \\
\end{tabular}
} 
\vspace{-12pt}
\end{table}

%% file: tex/conclusion.tex
\section{Conclusion}
\label{sec:conclusion}
We have presented a new approach for creating computationally efficient CNN models with sparse kernels. Our approach benefits from the base kernels that are sparse and complementary for extracting distinct and discriminative features. This enables us to combine base kernels to represent more complex kernels to enrich feature representation. The combination is simple but efficient to improve performance, and the presented fusion approaches are almost cost-free. Furthermore, with those deterministic sparse kernels, we achieved higher speed up in practice easily which is not completed by others.
We have integrated our approach into various existing CNN models and conducted extensive experiments to demonstrate the effectiveness of the approach for multiple tasks, like object classification, object localization and object detection, which validates the adaptability of the proposed method is effective. For classification and localization, \scFusion always outperforms the baselines and related work for all networks; on the other hand, our detector is 3$\times$ faster and 12.4$\times$ smaller than the baseline with competitive performance.

%% file: tex/supplementary.tex







\noindent
{\LARGE
\textbf{Appendix} \\
}

\section{Object Classification}

\begin{figure*}[bth]
    \centering
    \includegraphics[width=0.95\linewidth]{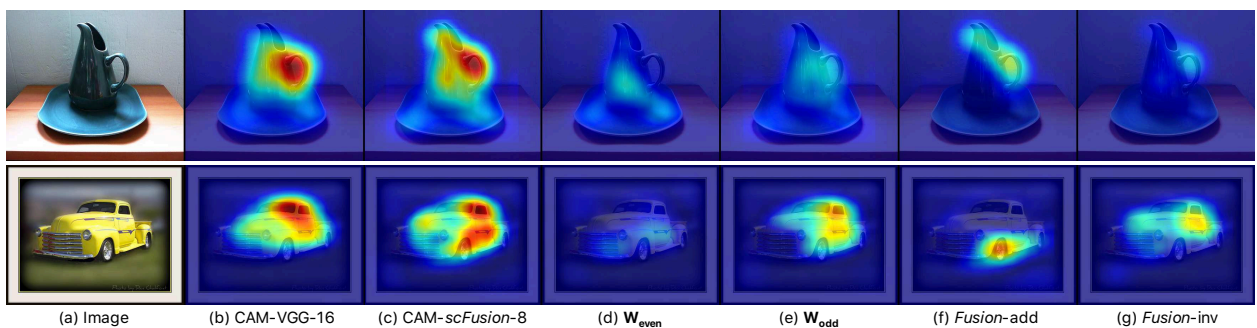}
    \caption{Class attention map from each component. (a) \scFusion-8 of VGG-16. (b) the baseline VGG-16-GAP, (c) the \scFusion-8 and (d)-(g) denotes the CAM maps from each component of \scFusion-8.}
    \label{fig:vgg16-individal-vis}
\end{figure*}

To facilitate reproducing our work, we provide details about the settings we used to train various models in our experiments. While the experimental settings can be different from datasets and models, we have applied identical hyper-parameters for the same network architecture for fair comparison.
All networks were trained with the momentum optimizer and the momentum is set to 0.9. 

\subsection{CIFAR-10}

For ResNet-32, ResNet-164 and DenseNet-40, the learning rate starts from 0.1 and then drops 10$\times$ at 50\% and 75\% epochs. We trained 185 epochs for both ResNets and 300 epochs for DenseNet. The batch size is set to 128 for these three networks. During the training, 4 pixels are padded on four directions and then a 32$\times$32 patch is cropped to increase its translation invariance. In Channel-wise mean and standard deviation normalization as well as random horizontal flipping are applied for data augmentation. In the testing phase, only the channel-wise mean and standard deviation normalization is used.

\subsection{ImageNet}
For AlexNet, we trained 90 epochs with a batch size of 256. The initial learning rate is 0.01, and drops 10$\times$ at the 20-th, 40-th, 60-th and 80-th epochs, respectively. We augmented data with horizontal flipping and pixel-wise mean subtraction.

For VGG-16 and ResNet-34, we trained 100 epochs with a batch size 256. 
The initial learning rate is 0.01 and 0.1 for VGG-16 and ResNet-34, respectively, and drops 10$\times$ at the 30-th, 60-th and 90-th epochs.
We applied scale augmentation in~\cite{Szegedy_CVPR_2015}, channel-wise mean and standard deviation normalization and random horizontal flipping. 

For the \scFusion-8 on VGG-16 with the deep compression proposed in \cite{Han_ICLR_2016}, we only fine-tuned the fully-connected layers.
We deleted the weights whose absolute values are below 0.0001, and then kept them as zeros during fine-tuning.
Afterwards, we trained 30 epochs with an initial learning rate of 0.001, which then drops 10$\times$ at the 20-th and 25-th epochs.

To measure the accuracy on the ImageNet validation dataset, we resized the shorter side of an image to 256 with its aspect ratio and then cropped a 224$\times$224 center patch for evaluation.

\section{Object Localization}

Following the methodology used in~\cite{Zhou_CVPR_2016}, for the VGG-16-based \scFusion model, we added another $3\times3$ convolutional layer with $1,024$ output channels after $conv5\_3$, and replaced the fully-connected layers by global average pooling, and then fine-tuning the model using ILSVRC-2012 training data. 
We then applied the same algorithm and parameters as used in~\cite{Zhou_CVPR_2016} to localize an object based on the class attention map (CAM) for each of the top-5 predictions. 
Shown in Table~\ref{table:localization-error-sup} are the top-1 and top-5 localization accuracy on the validation set of ILSVRC-2014.

To understand why our approach is effective at object localization, we take a closer look at the two examples in Fig.~\ref{fig:vgg16-individal-vis}, which illustrates the contribution of each component of the \scFusion module ($\mathbf{W_{even}}$, $\mathbf{W_{add}}$, fusion-add, fusion-inv). 
From them, we can learn a) Fusion operation, either addition or inverse, learns similar salient parts of an object, indicating they provide strong approximation of full kernels;  b) $\mathbf{W_{even}}$ and $\mathbf{W_{odd}}$ activate different parts of an object, indicating they contribute supplementally to object classification; and c) Combing all of the elements leads to a stronger activation than the original model.

\begin{table}[t]
\centering
\caption{Localization accuracy on the ILSVRC-2014 validation set.}
\label{table:localization-error-sup}
\resizebox{\linewidth}{!}{
\begin{tabular}{l|c|cc}
\toprule
Network                & Top-1 Acc. & Top-5 Acc. \\
\midrule
VGG-16-\scFusion-4-GAP & \textbf{48.12\%} & \textbf{59.96\%} \\
VGG-16-\scFusion-8-GAP &  46.11\% & 58.38\% \\
VGG-16-GAP~\cite{Zhou_CVPR_2016} & 42.80\% & 54.86\% \\
GoogLeNet~\cite{Zhou_CVPR_2016} &  43.60\%   &   57.00\% \\
\bottomrule
\end{tabular}
} 
\end{table}

%% file: main.bbl
\begin{thebibliography}{10}\itemsep=-1pt

\bibitem{Everingham2010}
M.~Everingham, L.~Van~Gool, C.~K.~I. Williams, et~al.
\newblock {The Pascal Visual Object Classes (VOC) Challenge}.
\newblock {\em IJCV}, 88(2):303--338, June 2010.

\bibitem{Fan_BMVC_2017}
Q.~Fan, C.-F. Chen, and G.~G. Lee.
\newblock A sparse deep feature representation for object detection from
  wearable cameras.
\newblock In {\em BMVC}, 2017.

\bibitem{py-faster-rcnn}
R.~Girshick.
\newblock py-faster-rcnn.
\newblock \url{https://github.com/rbgirshick/py-faster-rcnn}, 2015.

\bibitem{Han_ISCA_2016}
S.~Han, X.~Liu, H.~Mao, et~al.
\newblock {EIE: Efficient Inference Engine on Compressed Deep Neural Network}.
\newblock In {\em ISCA}, 2016.

\bibitem{Han_ICLR_2016}
S.~Han, H.~Mao, and W.~J. Dally.
\newblock {Deep Compression: Compressing Deep Neural Network with Pruning,
  Trained Quantization and Huffman Coding}.
\newblock In {\em ICLR}, 2016.

\bibitem{He_CVPR_2016}
K.~He, X.~Zhang, S.~Ren, et~al.
\newblock {Deep Residual Learning for Image Recognition}.
\newblock In {\em CVPR}, June 2016.

\bibitem{He_2016_ECCV}
K.~He, X.~Zhang, S.~Ren, et~al.
\newblock {Identity Mappings in Deep Residual Networks}.
\newblock In {\em ECCV}, 2016.

\bibitem{Hinton_arXiv_2015}
G.~Hinton, O.~Vinyals, and J.~Dean.
\newblock {Distilling the knowledge in a neural network}.
\newblock {\em arXiv}, 2015.

\bibitem{Huang_2017_CVPR}
G.~Huang, Z.~Liu, L.~van~der Maaten, et~al.
\newblock {Densely Connected Convolutional Networks}.
\newblock In {\em CVPR}, July 2017.

\bibitem{Iandola_ICLR_2017}
F.~N. Iandola, M.~W. Moskewicz, K.~Ashraf, et~al.
\newblock {SqueezeNet: AlexNet-level accuracy with 50x fewer parameters and
  $<$0.5MB model size}.
\newblock In {\em ICLR}, 2016.

\bibitem{Ioannou_ICLR_2016}
Y.~Ioannou, D.~P. Robertson, J.~Shotton, et~al.
\newblock {Training CNNs with Low-Rank Filters for Efficient Image
  Classification}.
\newblock In {\em ICLR}, 2016.

\bibitem{Jaderberg_BMVC_2014}
M.~Jaderberg, A.~Vedaldi, and A.~Zisserman.
\newblock Speeding up convolutional neural networks with low rank expansions.
\newblock In {\em BMVC}, 2014.

\bibitem{Jia_ACMMM_2014}
Y.~Jia, E.~Shelhamer, J.~Donahue, et~al.
\newblock {Caffe: Convolutional Architecture for Fast Feature Embedding}.
\newblock In {\em ACMMM}, pages 675--678, 2014.

\bibitem{Simonyan_ICLRWS_2014}
A.~V. K.~Simonyan and A.~Zisserman.
\newblock {Deep in- side convolutional networks: Visualising image classifica-
  tion models and saliency maps}.
\newblock In {\em ICLRWS}, 2014.

\bibitem{cifar10}
A.~Krizhevsky and G.~Hinton.
\newblock Learning multiple layers of features from tiny images.
\newblock Technical report, University of Toronto, 2009.

\bibitem{Krizhevsky_NIPS_2012}
A.~Krizhevsky, I.~Sutskever, and G.~E. Hinton.
\newblock {Imagenet classification with deep convolutional neural networks}.
\newblock In {\em NIPS}, 2012.

\bibitem{Lavin_CVPR_2016}
A.~Lavin and S.~Gray.
\newblock Fast algorithms for convolutional neural networks.
\newblock In {\em CVPR}, 2016.

\bibitem{Hao_ICLR_2017}
H.~Li, A.~Kadav, I.~Durdanovic, et~al.
\newblock {Pruning Filters for Efficient ConvNets}.
\newblock In {\em ICLRW}, pages 1--13, Mar. 2017.

\bibitem{Lin_2014_ECCV}
T.-Y. Lin, M.~Maire, S.~Belongie, et~al.
\newblock {Microsoft COCO: Common Objects in Context}.
\newblock In {\em ECCV}, pages 740--755, 2014.

\bibitem{Liu_CVPR_2015}
B.~Liu, M.~Wang, H.~Foroosh, M.~Tappen, and M.~Pensky.
\newblock Sparse convolutional neural networks.
\newblock In {\em CVPR}, 2015.

\bibitem{Liu_ICCV_2017}
Z.~Liu, J.~Li, Z.~Shen, et~al.
\newblock {Learning Efficient Convolutional Networks Through Network Slimming}.
\newblock In {\em ICCV}, Oct. 2017.

\bibitem{Luo_ICCV_2017}
J.-H. Luo, J.~Wu, and W.~Lin.
\newblock {ThiNet: A Filter Level Pruning Method for Deep Neural Network
  Compression}.
\newblock In {\em ICCV}, Oct. 2017.

\bibitem{ShuffleNetV2}
N.~Ma, X.~Zhang, H.-T. Zheng, and J.~Sun.
\newblock {ShuffleNet V2: Practical Guidelines for Efficient CNN Architecture
  Design}.
\newblock In {\em ECCV}, 2018.

\bibitem{Molchanov_ICLR_2017}
P.~Molchanov, S.~Tyree, T.~Karras, et~al.
\newblock {Pruning Convolutional Neural Networks for Resource Efficient
  Transfer Learning}.
\newblock In {\em ICLR}, 2017.

\bibitem{Peng_ECCV_2018}
B.~Peng, W.~Tan, Z.~Li, et~al.
\newblock {Extreme Network Compression via Filter Group Approximation}.
\newblock In {\em ECCV}, 2018.

\bibitem{Ren_TPAMI_2017}
S.~Ren, K.~He, R.~Girshick, and J.~Sun.
\newblock {Faster R-CNN: Towards Real-Time Object Detection with Region
  Proposal Networks}.
\newblock {\em TPAMI}, 39(6):1137--1149, June 2017.

\bibitem{ILSVRC15}
O.~Russakovsky, J.~Deng, H.~Su, et~al.
\newblock {ImageNet Large Scale Visual Recognition Challenge}.
\newblock {\em IJCV}, 115(3):211--252, 2015.

\bibitem{MobileNetV2}
M.~Sandler, A.~Howard, M.~Zhu, et~al.
\newblock {MobileNetV2: Inverted Residuals and Linear Bottlenecks}.
\newblock In {\em CVPR}, June 2018.

\bibitem{Shang_ICML_2016}
W.~Shang, K.~Sohn, D.~Almeida, et~al.
\newblock {Understanding and Improving Convolutional Neural Networks via
  Concatenated Rectified Linear Units}.
\newblock In {\em ICML}, 2016.

\bibitem{Simonyan_ICLR_2015}
K.~Simonyan and A.~Zisserman.
\newblock {Very Deep Convolutional Networks for Large-Scale Image Recognition}.
\newblock In {\em ICLR}, 2015.

\bibitem{Sun_ECCV_2016}
Z.~Sun, M.~Ozay, and T.~Okatani.
\newblock {Design of Kernels in Convolutional Neural Networks for Image
  Classification}.
\newblock In {\em ECCV}, 2016.

\bibitem{Szegedy_CVPR_2015}
C.~Szegedy, W.~Liu, Y.~Jia, et~al.
\newblock {Going deeper with convolutions}.
\newblock In {\em CVPR}, pages 1--9, 2015.

\bibitem{Szegedy_CVPR_2016}
C.~Szegedy, V.~Vanhoucke, S.~Ioffe, et~al.
\newblock Rethinking the inception architecture for computer vision.
\newblock In {\em CVPR}, 2016.

\bibitem{Wen_NIPS_2016}
W.~Wen, C.~Wu, Y.~Wang, et~al.
\newblock {Learning Structured Sparsity in Deep Neural Networks}.
\newblock In {\em NIPS}, 2016.

\bibitem{Yu_2017_CVPR}
F.~Yu, V.~Koltun, and T.~Funkhouser.
\newblock {Dilated Residual Networks}.
\newblock In {\em CVPR}, July 2017.

\bibitem{NISP}
R.~Yu, A.~Li, C.-F. Chen, et~al.
\newblock {NISP: Pruning Networks Using Neuron Importance Score Propagation}.
\newblock In {\em CVPR}, June 2018.

\bibitem{Yu_CVPR_2017}
X.~Yu, T.~Liu, X.~Wang, et~al.
\newblock {On Compressing Deep Models by Low Rank and Sparse Decomposition}.
\newblock In {\em CVPR}, July 2017.

\bibitem{Zhang_CVPR_2016}
X.~Zhang, J.~Zou, X.~Ming, et~al.
\newblock {Efficient and Accurate Approximations of Nonlinear Convolutional
  Networks}.
\newblock In {\em CVPR}, 2015.

\bibitem{Zhou_CVPR_2016}
B.~Zhou, A.~Khosla, A.~Lapedriza, et~al.
\newblock {Learning Deep Features for Discriminative Localization}.
\newblock In {\em CVPR}, June 2016.

\end{thebibliography}
